# Improving Traffic Flow Predictions with SGCN-LSTM: A Hybrid Model for Spatial and Temporal Dependencies

Alexandru T. Cismaru[1†]

[1]*Homestead High School, 21370 Homestead Rd, Cupertino, CA 95014*
[†]*Corresponding Author: 1alexcis@gmail.com*

Large amounts of traffic can lead to negative effects such as increased car accidents, air pollution, and significant time wasted. Understanding traffic speeds on any given road segment can be highly beneficial for traffic management strategists seeking to reduce congestion. While recent studies have primarily focused on modeling spatial dependencies by using graph convolutional networks (GCNs) over fixed weighted graphs, the relationships between nodes are often more complex, with edges that interact dynamically. This paper addresses both the temporal patterns in traffic data and the intricate spatial dependencies by introducing the Signal-Enhanced Graph Convolutional Network Long Short Term Memory (SGCN-LSTM) model for predicting traffic speeds across road networks. Extensive experiments on the PEMS-BAY road network traffic dataset demonstrate the SGCN-LSTM model's effectiveness, yielding significant improvements in Mean Absolute Error (MAE), Root Mean Square Error (RMSE), and Mean Absolute Percentage Error (MAPE) compared to benchmark models on the same dataset.

## 1. INTRODUCTION

Road traffic accidents are the leading cause of death among young people and a major global mortality factor, causing 1.24 million deaths each year [1]. Studies indicate that the risk of fatal injury rises by 28% during afternoon rush hours and by 36% in the morning rush hour [2], showing that heavy traffic is a significant cause of car crashes. Understanding traffic volume and speed on specific road segments can aid traffic control systems [3]. Improved traffic control and route recommendation systems can help ease congestion, reducing accidents, air pollution, and time wasted on the road [4]. However, key traffic metrics are not always available, limiting these systems' effectiveness. Accurately predicting traffic volume is, therefore, essential for generating reliable data for traffic control systems.

Machine learning (ML) models have been widely applied to predict traffic flows using traffic data. Yu et al. (2017) introduced a model that incorporates spatial and temporal dependencies in traffic data through Spatio-Temporal Graph Convolutional Networks (STGCN) [5]. This model achieved a Mean Absolute Error (MAE) of 3.78 on the BJER4 dataset, aggregated over 15 minutes, outperforming six other models. However, its high computational demands in spatio-temporal graph convolutions limit scalability for larger datasets. Yu (2022) developed a graph construction scheme for traffic prediction using graph adjacency learning and attention mechanisms to dynamically adjust sensor correlations, reducing prediction error (RMSE) by 5.60% across three real-world datasets. Additional improvements of 2.28% to 5.13% were seen using time-varying graphs [6]. Despite these improvements, dynamically updating large matrices can limit prediction efficiency on a metropolitan scale.

Bai et al. (2021) introduced the A3T-GCN model, which combines Graph Convolutional Networks (GCNs) with Gated Recurrent Units (GRUs) and an attention mechanism to capture spatial dependencies in road networks and temporal dynamics in traffic flow [7]. Their model achieved reductions in RMSE by up to 46.15% and increased accuracy by 10.37% on the SZ-taxi and Los-loops datasets compared to baseline models. However, integrating GCNs and GRUs with an attention mechanism can complicate the balance between spatial and temporal features, potentially affecting performance when one type of feature is dominant.

This paper introduces the SGCN-LSTM model, a hybrid of Signal-Enhanced Graph Convolutional Network (SGCN) and Long Short-Term Memory (LSTM) networks, designed to address the limitations of previous models. Unlike Yu et al.'s (2017) model, the SGCN-LSTM reduces computational demands by efficiently integrating spatial and temporal dependencies. It also employs graph adjacency learning and attention to dynamically establish node-level correlations, avoiding the static adjacency matrices used by Yu (2022). Finally, by combining the spatial feature handling of GCNs with the temporal sequence modeling of LSTMs, SGCN-LSTM effectively balances spatial and temporal dependencies, addressing the integration challenge faced by Bai et al.'s (2021) A3T-GCN model.

## 2. METHODOLOGY

*2.1 Dataset*

PEMS-BAY is a widely recognized public traffic dataset containing temporal traffic data for road segments in the Bay Area. The dataset provides time-series data specifically on traffic speed, which allows us to predict this key measure of traffic flow and manage congestion. Each entry includes features like traffic speed on various road segments over specific time intervals and road network information such as edge indices and edge weights.

*2.2 Graph Preprocessing and Feature Embedding*

Using the existing graph structure in the PEMS-BAY dataset, we preprocess the data to retain essential traffic features and spatial relationships between road segments. Each node represents a road segment, while edges capture the connectivity between segments, with edge weights reflecting spatial correlations. To maintain consistent indexing and retain important edge attributes throughout the dataset, we implement a custom data structure that incorporates edge weights at each time step. This preprocessing approach ensures that each data snapshot captures both temporal dynamics and spatial dependencies inherent within the graph structure.

*2.3 Graph to Signal Conversion*

To support temporal learning, we transformed the PEMS-BAY dataset into a time-series format suitable for short-term traffic pattern analysis. We defined a sequence length of N=1, instructing the model to focus on the initial 5-minute interval of each sequence in the dataset. This setup enables us to evaluate short-term traffic patterns efficiently, even with limited processing power. We standardized features and targets to ensure values are zero-centered with unit variance. This scaling step aids in stabilizing model training and enhances convergence by reducing sensitivity to variable magnitudes in the dataset.

*2.4 The SGCN-LSTM Model Architecture*

To predict traffic speeds using a Graph Neural Network (GNN) that captures both spatial and temporal characteristics of traffic data, we designed the Signal-Enhanced Graph Convolutional Network - Long Short Term Memory (SGCN-LSTM) model. This model architecture combines a Graph Convolutional Network (GCN) for spatial dependencies between road segments

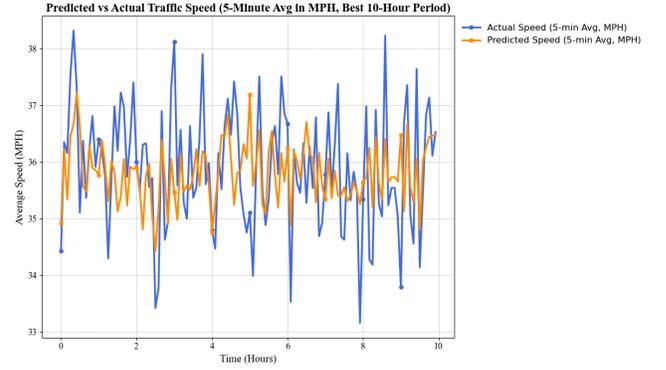

FIG. 1. This figure shows the average traffic speed (in MPH) over 5-minute intervals for a 10-hour period, comparing the actual recorded speeds (in blue) against the model's predicted speeds (in orange). The x-axis represents the time in hours, and the y-axis denotes the average speed in miles per hour. The figure demonstrates the model's ability to follow the general trend of actual speed variations, with the predicted line closely mirroring the overall pattern of the actual data. However, discrepancies between the two lines suggest that the model may be less accurate in predicting outlier speeds, highlighting areas where prediction performance could vary across different traffic conditions.

with a Long Short Term Memory (LSTM) network for capturing temporal patterns in traffic flow data.

In our architecture, the GCN component consists of two graph convolutional layers, each operating on the node features and edge weights to capture spatial relationships between connected road segments. The first GCN layer transforms input features into a hidden representation, followed by a ReLU activation function. This transformation is represented as $H^{(1)} = ReLU(D^{-1/2}AD^{-1/2}XW^{(0)})$, where $A$ denotes the adjacency matrix, $D$ is the degree matrix, $X$ is the feature matrix, and $W^{(0)}$ is the weight matrix of the first GCN layer. The output of the first GCN layer, $H^{(1)}$, is passed through a second GCN layer to produce $H^{(2)}$, calculated by $H^{(2)} = ReLU(D^{-1/2}AD^{-1/2}XW^{(1)})$, where $W^{(1)}$ represents the weight matrix for the second GCN layer. This process results in a spatially enriched feature representation for each node, which is then reshaped to serve as input to the LSTM component.

The LSTM component, designed to capture temporal dependencies, consists of a single LSTM layer that processes sequences of spatially enriched features from the GCN while maintaining information across time steps through its gating mechanisms: input, forget, and output gates. The LSTM's output from the last time step is passed directly to a fully connected linear layer to predict the final traffic speed for each node. This prediction is formulated as $Y_{pred} = Linear(H_t)$, where $H_t$ is the hidden state of the last LSTM layer, and *Linear* represents the transformation layer applied to produce the final prediction.

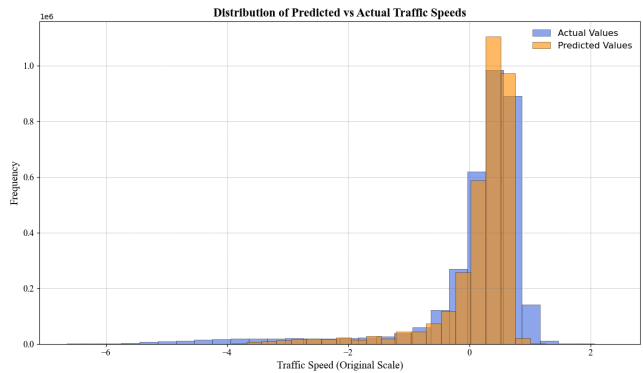

FIG. 2. This plot shows the distribution of the SGCN-LSTM's predicted vs. actual traffic speeds on the PEMS-BAY dataset. The plot shows the model's capability to closely follow the general trend. However, outliers within the data may often lead to predictions that are either higher or lower than the actual values.

*2.5 Training and Validation*

We trained the SGCN-LSTM model on preprocessed PEMS-BAY data, splitting the dataset into training (80%) and test (20%) sets to assess model performance. The training process used a combined loss function that balanced MAE and MSE with a weight of α=0.7. This approach leveraged MAE for interpretability while using MSE to penalize larger errors, ensuring robust accuracy across both typical and outlier traffic speeds.

We employed the Adam optimizer with a learning rate of 5e-4. Given the model's complexity and to prevent overfitting, we employed early stopping with a patience threshold of 5 epochs, halting training when validation loss failed to improve. The training loop processed mini-batches of 128 samples, and weights were updated using backpropagation. Additionally, gradient clipping was applied to stabilize training, especially when handling large datasets.

## 3. RESULTS

*3.1 Evaluation of Traffic Speed Prediction on PEMS-BAY*

To assess the SGCN-LSTM model's performance, we tested its ability to predict traffic speeds on the PEMS-BAY dataset, specifically for short-term traffic speed forecasting over 5-minute intervals. The evaluation was conducted by comparing the predicted speeds with the actual recorded speeds on each road segment within the dataset.

The results demonstrate that the SGCN-LSTM model achieves high accuracy over the specified time period, with a Mean Absolute Error (MAE) of 0.4347, indicating an average deviation of 0.4347 units from the actual values. This relatively low MAE suggests the model effectively captures and predicts slight fluctuations in traffic speeds over short intervals. Additionally, the model achieved a Mean Squared Error (MSE) of 0.6314 and a Root Mean Squared Error (RMSE) of 0.7946, highlighting that most predictions are closely aligned with the actual values, with only a few significant deviations.

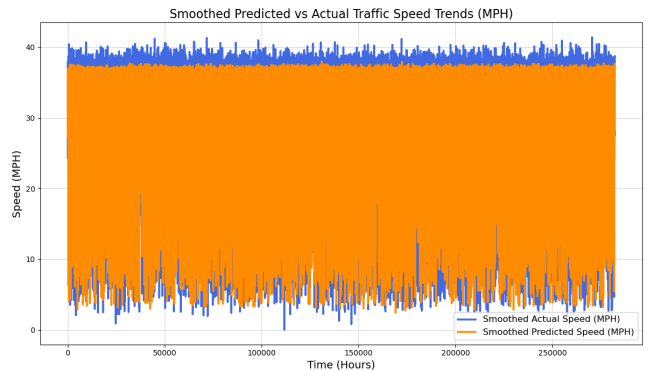

FIG. 3. This plot shows the range of the SGCN-LSTM's predictions (in orange) compared with the actual speed values (in blue) of the entire PEMS-BAY dataset. Although some small fluctuations remain, especially in capturing the higher-speed outliers, the model captures the appropriate range of the actual data over an extended time period, demonstrating that it performs well in capturing the general trend of traffic speeds.

| *Model Name* | *MAE* | *MSE* | *RMSE* |
|---|---|---|---|
| SGCN-LSTM | **0.4347** | **0.6314** | **0.7946** |
| A3T-GCN | 0.5743 | 0.9603 | 0.9799 |
| STGCN | 0.6175 | 1.0059 | 1.0029 |

Table 1. Comparison of MAE, MSE, and RMSE metrics for the SGCN-LSTM, A3T-GCN, and STGCN models.

These metrics confirm that the SGCN-LSTM model is effective at short-term traffic speed prediction.

*3.2 Comparison with A3T-GCN and STGCN Models*

As shown in Table 1, the SGCN-LSTM model outperforms the other models in terms of MAE, MSE, and RMSE.

The A3T-GCN and STGCN models both recorded higher MAE values of 0.5743 and 0.6175, respectively, indicating that their predictions deviated more on average from actual traffic speeds. This suggests that SGCN-LSTM has a better ability to capture finer variations in traffic patterns.

The MSE for A3T-GCN and STGCN was also higher, at 0.9603 and 1.0059 respectively, showing that both models struggled more with large prediction errors. In contrast, the lower MSE of SGCN-LSTM reflects its stronger capacity to minimize significant deviations across predictions.

Similarly, the RMSE values of A3T-GCN (0.9799) and STGCN (1.0029) indicate a tendency for greater prediction fluctuations compared to the SGCN-LSTM. The consistently lower error metrics across MAE, MSE, and RMSE confirm SGCN-LSTM's superior performance and reliability for short-term traffic speed forecasting on the PEMS-BAY dataset.

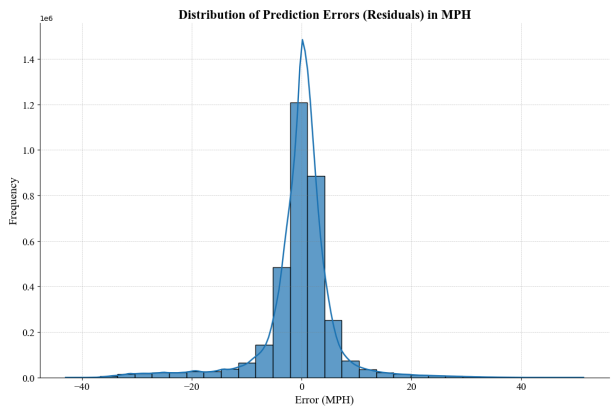

FIG. 4. This histogram shows the distribution of residuals, or prediction errors, in the SGCN-LSTM model's traffic speed predictions. The majority of errors are centered around zero, indicating that most predictions closely match the actual values. The distribution is relatively narrow with a slight spread to either side, suggesting that while the model is accurate in general, it does exhibit occasional deviations, both positive and negative. The residuals are concentrated around zero, demonstrating the model's strength in capturing the general trend of traffic speeds with limited significant errors.

## 4. DISCUSSION

*4.1 Challenges and Limitations*

One significant challenge in developing the SGCN-LSTM model is the quality and availability of traffic data. Traffic speed data can often be inconsistent, noisy, or missing, which can negatively impact model accuracy. Additionally, acquiring comprehensive datasets over extended periods or across diverse regions can be difficult. Our approach was designed with computational efficiency in mind by converting undirected graphs to temporal signal formats; however, training still proved resource-intensive with larger datasets. All computational tasks described in this paper were executed on an Apple M2 processor, which extended the training duration due to hardware constraints. Given our processing limitations, we focused on training the SGCN-LSTM model using only the 5-minute interval data from the PEMS-BAY dataset, as training and testing across larger time intervals was infeasible.

*4.2 Future Work*

We aim to further evaluate the SGCN-LSTM's capabilities by testing its performance on a wider range of time intervals within the PEMS-BAY dataset, such as 15-minute, 30-minute, and 60-minute intervals. We also hope to expand our analysis to other traffic flow metrics, such as traffic volume, to assess the model's generalization capabilities.

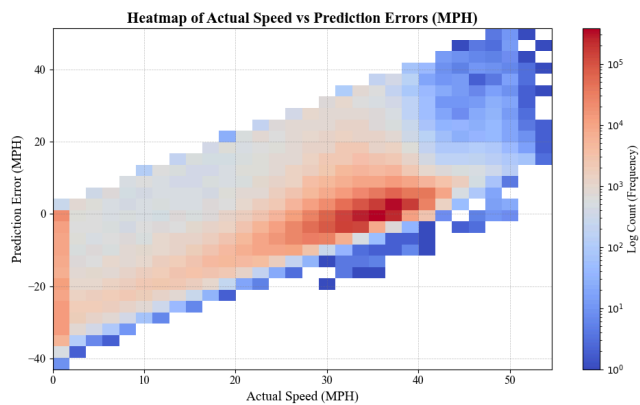

FIG. 5. This heatmap visualizes the distribution of prediction errors in relation to actual traffic speeds, with color intensity representing the log count of occurrences. Most errors are clustered around zero when speeds are moderate (20–30 MPH), indicating that the model performs well under typical traffic conditions. However, there is a trend of increased error magnitude at lower and higher speeds, suggesting that the model may struggle with extreme traffic scenarios, either when traffic is very slow or very fast. The concentration of errors finds itself being near zero, showing the model's limited errors in predictions on the PEMS-BAY dataset.

## 5. CONCLUSION

The SGCN-LSTM introduces a novel approach to predicting traffic flows, such as car volume. Our model's capabilities in predicting traffic volumes for future days and unknown road segments are beneficial for providing crucial information to traffic congestion management strategists. Whereas previous papers require intense computational power, such as Yu et al. (2017) and Yu (2022), who construct adjacency matrices to represent graph data, our paper is both computationally efficient and accurate by converting an undirected graph to a temporal signal format [5], [6]. We also mitigate the challenge of balancing spatial and temporal features found in other models, such as Bai et al. 's (2021) A3T-GCN, by integrating the spatial features of GCNs with the temporal capabilities of LSTMs. The code supporting our methodology, along with documentation, is publicly available on GitHub to facilitate reproducibility and further research on this topic by the community (https://github.com/atcis17/SGCN-LSTM-Predicting-Traffic-Speeds-on-PEMS-BAY). We will continue to improve upon the SGCN-LSTM by expanding the types of traffic flows it can predict and increasing its prediction accuracy through feature engineering and more descriptive data.

## ACKNOWLEDGEMENTS

We would like to express our sincere gratitude to Danial Alizadeh for his valuable guidance in clarifying publication rights for our project. His insights on the opportunities available in starting new projects provided us with the confidence and direction to pursue this research with the intention of sharing our findings more widely. We are grateful for his support and encouragement.